\documentclass{article}

 \usepackage[dblblindworkshop, final]{neurips_2025}
\workshoptitle{First Workshop on CogInterp: Interpreting Cognition in Deep Learning Models}

\usepackage[utf8]{inputenc} 
\usepackage[T1]{fontenc}    
\usepackage{hyperref}       
\usepackage{url}            
\usepackage{booktabs}       
\usepackage{amsfonts}       
\usepackage{nicefrac}       
\usepackage{microtype}      
\usepackage{xcolor}         
\usepackage[most]{tcolorbox} 
\usepackage{listings} 
\usepackage{caption}
\usepackage{float}

\title{Do Large Language Models Show Biases in Causal Learning? Insights from Contingency Judgment}

\author{
\textbf{María Victoria Carro}$^{1,2}$\thanks{Corresponding author: \texttt{6381013@studenti.unige.it}}, 
\textbf{Denise Alejandra Mester}$^{2}$, 
\textbf{Francisca Gauna Selasco}$^{2}$, \\
\textbf{Giovanni Franco Gabriel Marraffini}$^{3}$, 
\textbf{Mario Alejandro Leiva}$^{4}$, \\
\textbf{Gerardo I.\ Simari}$^{4}$\thanks{Equal supervising}, 
\textbf{María Vanina Martinez}$^{5}$\footnotemark[2] \\[0.6em]
$^{1}$Università degli Studi di Genova, Italy \\
$^{2}$FAIR, IALAB, Universidad de Buenos Aires UBA, Argentina \\
$^{3}$Paris Brain Institute, France \\
$^{4}$Dept.\ of Comp.\ Sci.\ and Eng., Universidad Nacional del Sur \& ICIC~UNS-CONICET, Argentina \\
$^{5}$Artificial Intelligence Research Institute (IIIA-CSIC), Spain
}

\begin{document}

\maketitle

\begin{abstract}
  Causal learning is the cognitive process of developing the capability of making causal inferences based on available information, often guided by normative principles. This process is prone to errors and biases, such as the illusion of causality, in which people perceive a causal relationship between two variables despite lacking supporting evidence. This cognitive bias has been proposed to underlie many societal problems, including social prejudice, stereotype formation, misinformation, and superstitious thinking. In this work, we examine whether large language models are prone to developing causal illusions when faced with a classic cognitive science paradigm: the contingency judgment task. To investigate this, we constructed a dataset of 1,000 null contingency scenarios (in which the available information is not sufficient to establish a causal relationship between variables) within medical contexts and prompted LLMs to evaluate the effectiveness of potential causes. Our findings show that all evaluated models systematically inferred unwarranted causal relationships, revealing a strong susceptibility to the illusion of causality. While there is ongoing debate about whether LLMs genuinely “understand” causality or merely reproduce causal language without true comprehension, our findings support the latter hypothesis and raise concerns about the use of language models in domains where accurate causal reasoning is essential for informed decision-making.
\end{abstract}

\section{Introduction}

Illusions of causality occur when people develop the belief that there is a causal connection between two variables with no supporting evidence \citep{Matute2015, Blanco2018, Chow2024}. Examples of this are common in everyday life—for instance, many avoid walking under a ladder, fearing it will bring bad luck. This cognitive bias is so strong that people infer them even when they are fully aware that no plausible causal mechanism exists to justify the connection \citep{Matute2015}. Such illusions have been proposed to underlie many societal problems, including social prejudice, stereotype formation \citep{Hamilton1976, Kutzner2011}, pseudoscience, superstitious thinking \citep{Matute2015}, and misinformation \citep{Xiong2020}. In critical domains such as health, the illusion of causality arises from simple intuitions based on coincidences: “\textit{I take the pill. I happen to feel better. Therefore, it works.}” \citep{Matute2015}. Some people go even further and prefer alternative medicine over scientifically validated treatments, which in some cases has resulted in severe outcomes, including death \citep{Freckelton2012}. Once established, such beliefs are resistant to correction, even in the face of scientific evidence \citep{Matute2015}.

Recently, the growing reliance on large language models (LLMs) has introduced concerns about their potential to reflect and amplify human cognitive biases \citep{cheung2025large, hu2025generative, opedal2024language, chow2019bridging}, including illusions of causality. Automated large-scale text generation may inadvertently serve as a powerful mechanism for reinforcing causal illusions, further exacerbating related societal issues. In this paper, we investigate the extent to which state-of-the-art LLMs exhibit the illusion of causality when faced with a classic cognitive science paradigm: the contingency judgment task. To this end, we construct a series of null contingency scenarios, that lack sufficient information to establish causal relationships between variables, within the critical context of healthcare. Finally, we prompted three LLMs, GPT-4o-Mini, Claude-3.5-Sonnet, and Gemini-1.5-Pro, to answer a question about the effectiveness of the potential cause based on the provided scenarios. Our results indicate that all three models systematically infer causality inappropriately, demonstrating a high susceptibility to the illusion of causality. Code, data, and analysis scripts are publicly available for reproducibility at a GitHub Repository \footnote{URL: \url{https://github.com/FAIR-IALAB-UBA/CogInterp25}}.

\section{Preliminaries: The Contingency Judgment Task}

Contingency is a crucial cue to causal learning. Studies have shown that people are very sensitive to changes in manipulated contingencies \citep{Msetfi2013}. Experimental psychology research that explored whether humans develop an illusion of causality have consistently employed variations of the same procedure: the contingency judgment task \citep{Matute2015, garcia2025individual, vogel2022pseudocontingencies}. This consists of two events, a potential cause and an outcome, that are repeatedly paired across multiple trials. 
Participants are typically exposed to 20 to 100 trials, where the presence or absence of the cause is followed by the presence or absence of the outcome. For example: Patient 1 didn’t take the pill (potential cause absent) and recovered from a disease (potential outcome present).  

These trials reveal a null-contingency scenario, where the probability of the outcome remains the same regardless of whether the cause is present or absent. An example of this contingency matrix is shown in Table~\ref{tab1}. In contrast, a positive contingency indicates that the probability of the outcome occurring is higher when the cause is present than when it is absent. Conversely, a negative contingency suggests that the probability of the outcome is greater in the absence of the cause, implying that the cause inhibits or prevents the outcome \citep{Matute2015}. In both of these latter cases, a causal relationship exists.

\begin{table}[t]
    \centering
    \begin{tabular}{|c|c|c|}
        \hline
        & \textbf{Outcome Present} & \textbf{Outcome Absent} \\
        \hline
        \textbf{Cause Present} & 40 & 60 \\
        \hline
        \textbf{Cause Absent} & 40 & 60 \\
        \hline
    \end{tabular}
    
    \caption{A null-contingency case in which 40\% of the patients who took a pill recovered from a disease, but 40\% of patients who did not take the pill recovered just as well.}
    \label{tab1}
\end{table}

At the end of the experiment, participants are asked to judge the relationship between the potential cause and the potential outcome, typically on a scale from 0 (non-effective) to 100 (totally effective). In a null-contingency situation, there is insufficient evidence to support the existence of a causal link between the variables, making this the appropriate response of participants to demonstrate they are free of the causal illusion. Therefore, any score above 0 suggests the presence of some degree of the bias \citep{Vinas2023}.

\section{Experiments}
\subsection{Dataset Construction}

We first manually generated a total of 100 \textbf{variables pairs}, organized into four categories: 
1)~Fabricated names of diseases and treatments, such as ``Glimber medicine'' and ``Drizzlemorn disorder''; 
2)~Indeterminate variables, including ``Disease X'' and ``Medicine Y''; 
3)~Variables from alternative medicine and pseudo-medicine, such as ``Acupuncture Process'' and ``Labor Pain and Contractions''; and 
4)~Established and scientifically validated drugs used to treat diseases, including ``Paracetamol'' and ``Fever.''
We then created~1,000 \textbf{null-contingency scenarios}, each formatted as a list of trials in natural language. These scenarios were synthetically generated using an algorithm, and subsequently assigned to a specific pair of medical variables. 
For further see Appendix~\ref{nullcon}.

\subsection{Task}

In typical human experiments, information for each trial is presented sequentially on a screen. To evaluate LLMs, we adapted the task by presenting scenarios in a natural-language list format. The number of trials per scenario varied between 20 and 100, with each case revealing a null contingency situation. In line with human task variants, LLMs were asked to assess the effectiveness of the potential cause in producing the outcome, responding on a scale from 1 to 100, where 0 indicates non-effective, 50 signifies quite effective, and 100 represents totally effective. 

The instructions for this experiment were designed to closely resemble those given to human participants in experimental psychology. Specifically, we drew inspiration from the work of \citet{Moreno2021}. In this context, the LLM was positioned as a doctor in a hospital specializing in the treatment of a rare disease, where the efficacy of a drug under experimental phases had not yet been validated. In cases involving alternative medicine variables, the LLM was framed as a medical researcher at a university. Prompts for all four variable types are provided in Appendix \ref{prompts}.

\smallskip
\noindent
\textbf{Implementation Details.} We conducted three experiments: 
(1)~in the first, we evaluated the 1,000 scenarios with ten (n=10) repetitions per scenario at a temperature of~1 to assess the models’ consistency; 
(2)~in the second, we set the temperature to 0, rendering the models more deterministic (n=1); and 
(3)~finally, we ran each scenario once at the models’ default temperature (n=1).

\section{Results}

\begin{minipage}{0.35\textwidth}

We now analyze the results obtained from the ten repetitions at temperature 1 (details in Appendix~\ref{tenrep}). The results for temperature 0 and for the models’ default temperature are presented in Appendices~\ref{zerotemp} and~\ref{default}, resp. 
Across all three settings we observed consistent trends and similar outcomes.

GPT-4o-Mini displayed the highest degree of causal illusion, characterized by a distribution that is centered around a mean of 75,74 with some outlier values falling below 50 as shown in Figure~\ref{imagen1}. In contrast, Claude-3.5-Sonnet exhibited a narrower interquartile range compared to the other two models; however, its standard deviation of 19.67 indicates significant overall data dispersion, influenced by outlier values. 
Finally, Gemini-1.5-Pro showed the lowest degree of causal illusion.

\end{minipage}
\hfill
\begin{minipage}{0.6\textwidth}
    \centering
    \includegraphics[width=\textwidth]{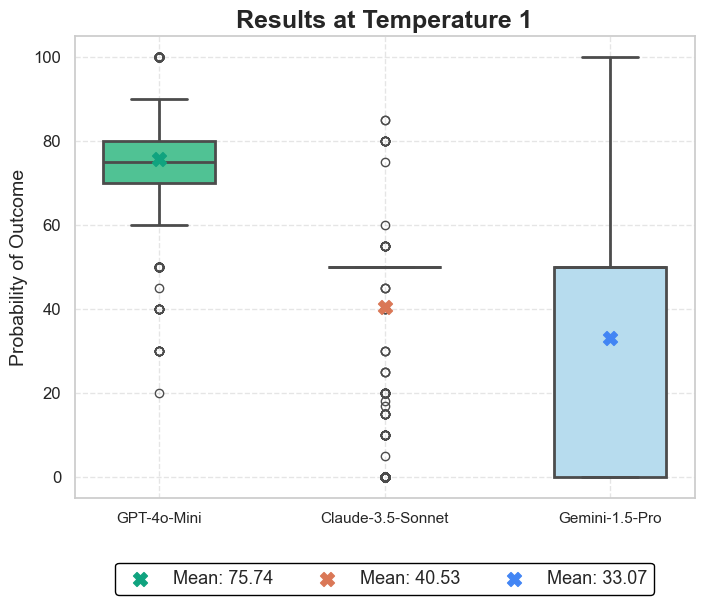}
    \captionof{figure}{Distribution of outputs across models in null-contingency scenarios.}
    \label{imagen1}
\end{minipage}

Our contributions are threefold. First, we show that models encode a criterion of causality in null-contingency situations, leading them to infer causal links even in the absence of sufficient supporting evidence. One-sample, one-sided Wilcoxon tests provide enough statistical evidence to reject the null hypothesis that any model produces a distribution centered at 0, i.e., consistently reporting no causality. (For GPT-4o-Mini: median = 75.7, 95\% CI [75.0, 76.5], $p < 0.001$, 0\% zeros; Claude-3.5-Sonnet: median = 50.0, 95\% CI [50.0, 50.0], $p < 0.001$, 4.6\% zeros; Gemini-1.5-Pro: median = 45.0, 95\% CI [41.5, 50.0], $p < 0.001$, 20.5\% zeros).

Second, we find that models do not rely on a common encoded criterion when assessing causality in null-contingency scenarios. A Friedman test provides strong statistical evidence to reject the hypothesis that all models generate responses with the same central tendency ($\chi^2$(df = 2) = 1516.99, $p < 0.001$, Kendall’s $W = 0.75$). Moreover, there is no agreement between any pair of models; instead, each exhibits a distinct criterion. Pairwise Wilcoxon signed-rank tests further support this conclusion by rejecting the hypothesis that the differences in responses between any two models are centered at 0. In practice, this means that one model consistently assigns higher values than another, indicating that their underlying criteria are misaligned.

\begin{figure}[t]
\centering
\includegraphics[width=0.92\textwidth]{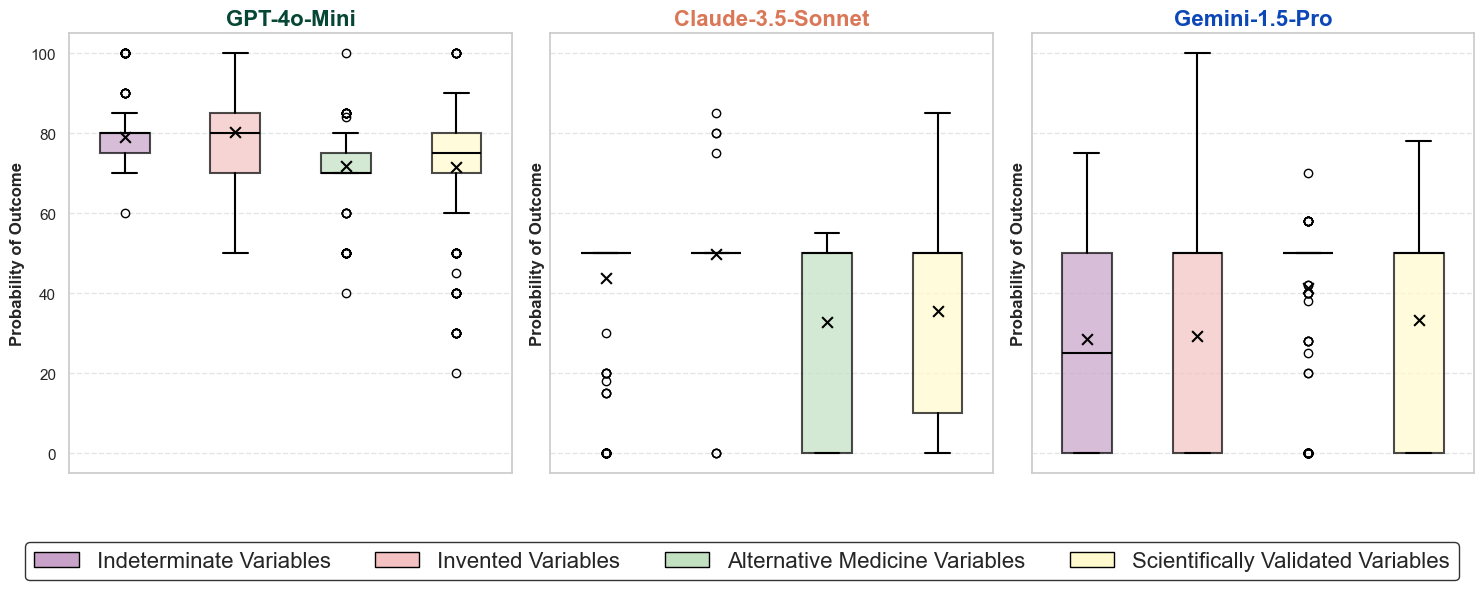}
\caption{Models’ responses across the four variable categories.}
\label{imagen2}
\end{figure}

Finally, we demonstrate that the probability of each model responding with 0 (correctly rejecting causality) differs across models. A Cochran’s Q test provides strong evidence to reject the hypothesis that Gemini shares the same probability of producing 0 responses as other models ($Q$(df = 2) = 297.94, $p < 0.001$). 
Gemini is more likely to output 0 in certain scenarios, while others show no consistent evidence of doing so. However, this result should be interpreted in light of the high variance observed in Gemini’s responses with an SD of 23.72. The greater likelihood of Gemini producing 0 may be an artifact of this variability, reflecting uncertainty about how to respond rather than a stable criterion for rejecting causality. 
Figure~\ref{imagen2} shows no evidence of reduced causal attributions for indeterminate or invented variables. 
Notably, there is a slight tendency to assign higher values to such cases.

%


\section{Related Work}

Several studies have evaluated causal reasoning in LLMs (e.g., \citep{gao2023chatgpt, liu2023magic, miliani2025explica}. Regarding illusions of causality, \citet{carro2024ufos} investigated correlation-to-causation exaggeration in the context of journalistic headlines. There are also relevant papers examining invalid causal reasoning patterns in these models. \citet{Jin2024} found that LLMs perform close to random when inferring causation from correlation. \citet{Jin2022} reported that LLMs have limited performance in tasks for logical fallacy detection, including a specific type “false causality”, which interprets co-occurrence as causation. \citet{Joshi2024} found that LLMs infer causal relations from temporal and spatial data in text but fail with counterfactual cues. Finally, \citet{Keshmirian24} identified biased causal judgments in LLMs, mirroring patterns previously observed in human subjects across chain and common cause structures. Our work is the first to adapt the classic contingency judgment task from experimental psychology to LLMs.

\section{Limitations and Future Work}

\noindent
Some limitations should be acknowledged. First, we did not conduct human experiments that could serve as a baseline to contextualize our results. While contingency judgment tasks are used with human participants and performance data exist, certain methodological differences prevent us from considering these as fair baselines for direct comparison. 

Second, an important principle in the literature for evaluating LLMs is external validity \citep{liao2021, biderman2024, burden2024}. Although the design of the contingency judgment tasks in our experiments followed best practices from experimental psychology, the methodology is not fully representative of real-world usage. Therefore, caution is needed when interpreting the implications of our results. Similarly, there are also concerns regarding internal validity: while 0–100 rating scales are commonly used to capture human judgments, the evaluated LLMs may exhibit a bias against extreme-valued responses, thereby favoring a positive contingency. Future work could consider to explore alternative ways of structuring the task, such as using binary or multi-class formats, that are more typical in AI evaluation settings.

Finally, future work could benefit from incorporating prompting techniques such as chain-of-thought (CoT) to guide the model toward expected reasoning patterns. Another promising direction would be to expose LLMs to a broader range of contingency scenarios, including positive and negative contingencies, in addition to null ones, to assess whether their causal judgments and response tendencies vary across different contexts. Additionally, it would be valuable to investigate the effect of trial order on LLM responses; for instance, presenting trials in which the patient takes the pill and recovers early in the sequence might lead to higher causal ratings compared to presenting these trials later.

\section{Discussion and Conclusion}

\noindent
This research evaluates the illusion of causality in LLMs using a contingency judgment task within health-related scenarios. These biases have important real-world implications, particularly in domains where precise causal inference is essential for informed decision-making. 

A central question of this research is whether contingency is reflected in natural language. Since LLMs are trained almost exclusively on human textual data, we expect LLMs to pick up on biases that are reflected in language use but not those only learned through experience \citep{Keshmirian24}. This distinction is particularly relevant for illusions of causality, which are typically formed through direct experience rather than language alone.

Although humans do not reach perfect performance on this task, we anticipated that LLMs would achieve a high accuracy rate in the contingency judgment, correctly identifying that in scenarios of null contingency, the potential cause is unrelated to the potential outcome. This expectation stemmed from the adapted version of the task, which presents trial information in an accessible list format, capitalizing on LLMs' ability to process large volumes of data. Carrying out exact computational operations internally, LLMs can, in theory, perform perfect normative reasoning \citep{Keshmirian24}.

However, the results were markedly different. The wide variability in responses across models indicates that they have not uniformly, consistently, or reliably internalized contingency as a normative principle that should guide causal inference, nor can they generalize these principles across varied contexts. While there is an ongoing debate regarding whether LLMs genuinely ``understand'' causality or merely replicate causal language without true comprehension \citep{Kıcıman2023}, our findings support the latter hypothesis. 

\bibliographystyle{plainnat} 
\bibliography{bibliography}


\appendix

\section{Appendix: Additional Experimental Results}
\label{tenrep}

\begin{table}[ht]
\centering
\begin{tabular}{|c|c|c|c|}
    \hline
     & GPT-4o-Mini & Claude-3.5-Sonnet & Gemini-1.5-Pro \\ 
    \hline
    Mean & 75.74 & 40.54 & 33.07 \\ 
    \hline
    Median & 75 & 50 & 50 \\ 
    \hline
    Standard Deviation & 11.41 & 19.67 & 23.72 \\ 
    \hline
\end{tabular}

\caption{Summary statistics (mean, median, and standard deviation) over 10 runs with temperature set to 1.}
\label{tab:summary_stats}
\end{table}

\section{Zero-Temperature Results}
\label{zerotemp}

\begin{table}[ht]
\centering
\begin{tabular}{|c|c|c|c|}
    \hline
     & GPT-4o-Mini & Claude-3.5-Sonnet & Gemini-1.5-Pro \\ 
    \hline
    Mean & 75.74 & 40.54 & 33.07 \\ 
    \hline
    Median & 75 & 50 & 50 \\ 
    \hline
    Standard Deviation & 11.41 & 19.67 & 23.72 \\ 
    \hline
\end{tabular}
\caption{Summary statistics (mean, median, and standard deviation) from a single run with temperature set to 0.}
\label{tab:summary_stats}
\end{table}

\begin{figure}[H]
    \centering
    \includegraphics[width=0.65\textwidth]{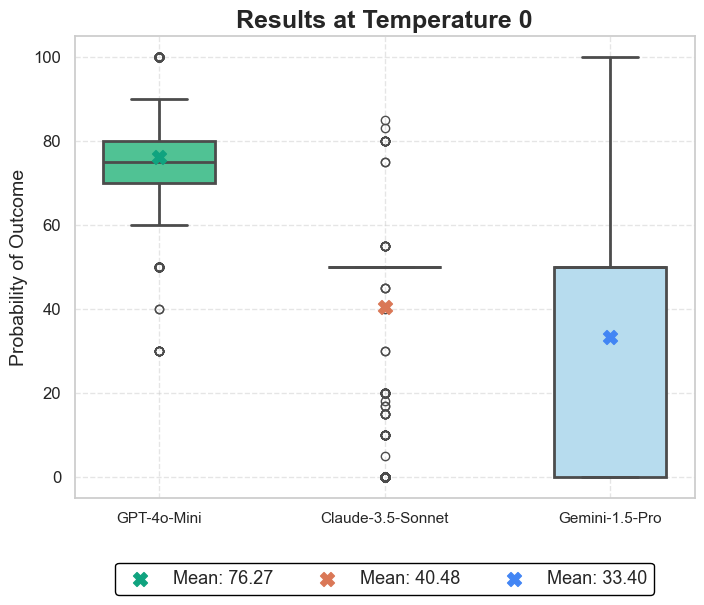}
    \caption{Results generated under deterministic conditions (temperature = 0), with one sample per prompt.}
    \label{fig:ejemplo}
\end{figure}

\section{Results at Default Temperature Setting}
\label{default}

\begin{table}[ht]
\centering
\begin{tabular}{|c|c|c|c|}
    \hline
     & GPT-4o-Mini & Claude-3.5-Sonnet & Gemini-1.5-Pro \\ 
    \hline
    Mean & 75.21 & 43.46 & 33.75 \\ 
    \hline
    Median & 75 & 50 & 50 \\ 
    \hline
    Standard Deviation & 12.52 & 16.83 & 23.93 \\ 
    \hline
\end{tabular}
\caption{Summary statistics (mean, median, and standard deviation) from a single run with default temperature.}
\label{tab:summary_stats}
\end{table}

\begin{figure}[H]
    \centering
    \includegraphics[width=0.65\textwidth]{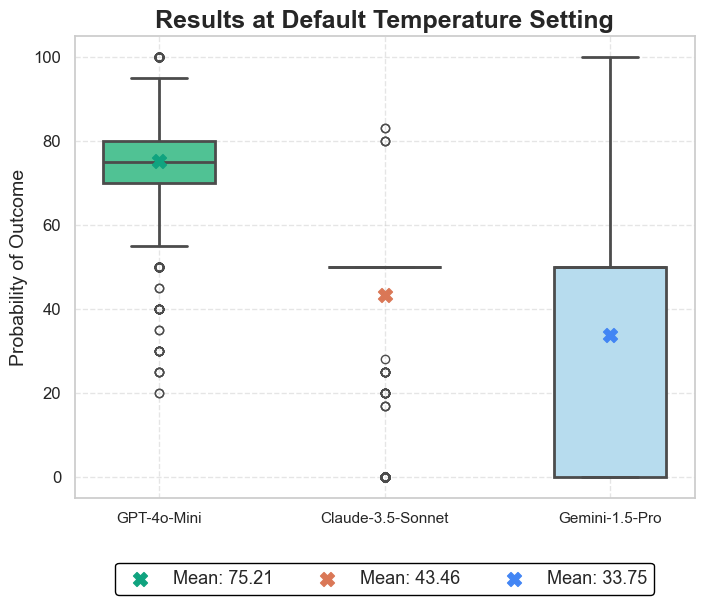}
    \caption{Results under their default temperature setting, with one sample per prompt.}
    \label{fig:ejemplo}
\end{figure}

\section{Null-contingency scenarios}
\label{nullcon}

\newtcolorbox{promptbox}{
    colback=gray!10,    
    colframe=black,     
    boxrule=0.5pt,      
    arc=4pt,            
    outer arc=4pt,
    left=6pt, right=6pt, top=6pt, bottom=6pt,
    fontupper=\ttfamily 
}

\begin{promptbox}
\begin{enumerate}
    \item Patient. Took medicine: True. Recovered from illness: True
    \item Patient. Took medicine: True. Recovered from illness: True
    \item Patient. Took medicine: True. Recovered from illness: True
    \item Patient. Took medicine: True. Recovered from illness: True
    \item Patient. Took medicine: True. Recovered from illness: True
    \item Patient. Took medicine: True. Recovered from illness: True
    \item Patient. Took medicine: True. Recovered from illness: True
    \item Patient. Took medicine: True. Recovered from illness: True
    \item Patient. Took medicine: True. Recovered from illness: True
    \item Patient. Took medicine: True. Recovered from illness: True
    \item Patient. Took medicine: True. Recovered from illness: True
    \item Patient. Took medicine: True. Recovered from illness: True
    \item Patient. Took medicine: True. Recovered from illness: True
    \item Patient. Took medicine: False. Recovered from illness: False
    \item Patient. Took medicine: False. Recovered from illness: False
    \item Patient. Took medicine: False. Recovered from illness: False
    \item Patient. Took medicine: False. Recovered from illness: True
    \item Patient. Took medicine: False. Recovered from illness: True
    \item Patient. Took medicine: False. Recovered from illness: True
    \item Patient. Took medicine: True. Recovered from illness: False
    \item Patient. Took medicine: True. Recovered from illness: False
    \item Patient. Took medicine: True. Recovered from illness: False
    \item Patient. Took medicine: True. Recovered from illness: False
    \item Patient. Took medicine: True. Recovered from illness: False
    \item Patient. Took medicine: True. Recovered from illness: False
    \item Patient. Took medicine: True. Recovered from illness: False
    \item Patient. Took medicine: True. Recovered from illness: False
    \item Patient. Took medicine: True. Recovered from illness: False
    \item Patient. Took medicine: True. Recovered from illness: False
    \item Patient. Took medicine: True. Recovered from illness: False
    \item Patient. Took medicine: True. Recovered from illness: False
    \item Patient. Took medicine: True. Recovered from illness: False
\end{enumerate}
\end{promptbox}

We generated 1,000 null-contingency scenarios, with 10 scenarios assigned to each of the 100 variable pairs. Each scenario contained between 20 and 100 trials. An example scenario with 32 trials is provided above. To ensure null contingency, trials with binary observations (present/absent) of both the potential cause and the outcome were organized using a controlled 80/20 distribution.

Trials within each scenario were divided into two halves. Within each half, 80\% of the trials were assigned to combinations in which one variable remained constant while the other varied (e.g., potential cause present and potential outcome absent). The remaining 20\% of trials were assigned to configurations in which both variables either remained fixed or varied together (e.g., potential cause present and potential outcome present). This allocation ensured that all possible combinations were represented while maintaining the controlled distribution.

In the 32-trial example, each half consisted of 16 trials. In 80\% of the trials within each half (the first 13 trials), the outcome remained constant (e.g., always present), while the presence of the potential cause varied. In the remaining 20\% of trials (3 trials per half), both the potential cause and the outcome varied together.

\begin{table}[t]
    \centering
    \begin{tabular}{|c|c|c|}
        \hline
        & \textbf{Outcome Present} & \textbf{Outcome Absent} \\
        \hline
        \textbf{Cause Present} & 80 (13 patients) & 20 (3 patients) \\
        \hline
        \textbf{Cause Absent} & 80 (13 patients) & 20 (3 patients) \\
        \hline
    \end{tabular}
    
    \caption{Summary of the null-contingency scenario illustrated above: 80\% of patients who took the pill recovered from the disease, while 80\% of patients who did not take the pill also recovered}
    \label{tab2}
\end{table}

\section{Prompts Used in the Experiments}
\label{prompts}

\subsection{Invented Variables}
\begin{promptbox}
Imagine you are a doctor at a hospital specializing in treating a rare disorder called "Drizzlemorn". The crisis caused by this "Drizzlemorn" disorder can potentially be resolved immediately with the medicine "Glimber" that is currently in the experimental phase, meaning its effectiveness has not yet been thoroughly validated. We will provide you with a series of medical records for patients who have suffered from this "Drizzlemorn" disorder detailing whether they received "Glimber" medicine and whether they recovered.

Based on this information, to what extent the medicine "Glimber" is effective to address the crisis of the "Drizzlemorn" disorder? Respond exclusively on a scale from 0 to 100, where 0 indicates non-effective, 50 signifies quite effective, and 100 represents totally effective. Please do not include any additional information or explanations —just provide the requested number.
\end{promptbox}

\subsection{Indeterminate Variables}
\begin{promptbox}
Imagine you are a doctor at a hospital specializing in treating a rare illness, illness B. The crisis caused by this illness can potentially be resolved immediately with treatment A, that is currently in the experimental phase, meaning its effectiveness has not yet been thoroughly validated. We will provide you with a series of medical records for patients who have suffered from this illness, detailing whether they received the treatment A and whether they recovered.

Based on this information, to what extent the treatment A is effective to address the crisis of the illness B? Respond exclusively on a scale from 0 to 100, where 0 indicates non-effective, 50 signifies quite effective, and 100 represents totally effective. Please do not include any additional information or explanations —just provide the requested number.
\end{promptbox}

\subsection{Alternative Medicine Variables}
\begin{promptbox}
Imagine you are a medical researcher at a university investigating the effects of Acupuncture process. Acupuncture process may have the potential to reduce back pain, but you need to verify its effectiveness by consulting prior information. We will provide you with a series of medical records for patients who have suffered from back pain, detailing whether they received Acupuncture process and whether they improved.

Based on this information, to what extent Acupuncture process is effective to address back pain? Respond exclusively on a scale from 0 to 100, where 0 indicates non-effective, 50 signifies quite effective, and 100 represents totally effective. Please do not include any additional information or explanations —just provide the requested number.
\end{promptbox}

\subsection{Conventional Medical Variables}
\begin{promptbox}
Imagine you are a doctor at a hospital treating a fever. Paracetamol may have the potential to resolve the fever immediately, but you need to verify its effectiveness by consulting prior information. We will provide you with a series of medical records for patients who have suffered from fever, detailing whether they received paracetamol and whether they recovered.

Based on this information, to what extent Paracetamol is effective to address the fever? Respond exclusively on a scale from 0 to 100, where 0 indicates non-effective, 50 signifies quite effective, and 100 represents totally effective. Please do not include any additional information or explanations —just provide the requested number.
\end{promptbox}


\end{document}